\documentclass[letterpaper]{article} 
\usepackage{aaai23}  
\usepackage{times}  
\usepackage{helvet}  
\usepackage{courier}  
\usepackage[hyphens]{url}  
\usepackage{graphicx} 
\usepackage{natbib}  
\usepackage{caption} 
\usepackage{algorithm}
\usepackage{algorithmic}

\nocopyright

\usepackage{newfloat}
\usepackage{listings}
\DeclareCaptionStyle{ruled}{labelfont=normalfont,labelsep=colon,strut=off} 
\floatstyle{ruled}
\newfloat{listing}{tb}{lst}{}
\floatname{listing}{Listing}
\title{ssFPN: Scale Sequence (S$^2$) Feature Based-Feature Pyramid Network\\for Object Detection}
\author{
    Hye-Jin Park\textsuperscript{\rm 1},
    Young-Ju Choi\textsuperscript{\rm 1},
    Young-Woon Lee\textsuperscript{\rm 2}, 
    Byung-Gyu Kim\textsuperscript{\rm 1}*
}
\affiliations{
    \textsuperscript{\rm 1}Sookmyung Women's University\\
    \textsuperscript{\rm 2}Sunmoon University\\

    \{hj.park, yj.choi, yw.lee\}@ivpl.sookmyung.ac.kr,
    bg.kim@sookmyung.ac.kr*
}

\usepackage{bibentry}

\begin{document}

\maketitle

\begin{abstract}

Feature Pyramid Network (FPN) has been an essential module for object detection models to consider various scales of an object. However, average precision (AP) on small objects is relatively lower than AP on medium and large objects. The reason is why the deeper layer of CNN causes information loss as feature extraction level. We propose a new scale sequence (S$^2$) feature extraction of FPN to strengthen feature information of small objects. We consider FPN structure as scale-space and extract scale sequence (S$^2$) feature by 3D convolution on the level axis of FPN. It is basically a scale-invariant feature and is built on high-resolution pyramid feature map for small objects. Furthermore, the proposed S$^2$ feature can be extended to most object detection models based on FPN. We demonstrate the proposed S$^2$ feature can improve the performance of both one-stage and two-stage detectors on MS COCO dataset. Based on the proposed S$^2$ feature, we achieve upto 1.3\% and 1.1\% of AP improvement for YOLOv4-P5 and YOLOv4-P6, respectively. For Faster R-CNN and Mask R-CNN, we observe upto 2.0\% and 1.6\% of AP improvement with the suggested S$^2$ feature, respectively.

\end{abstract}

\section{Introduction}

Object detection is an essential one of the fundamental tasks in computer vision. It has been widely used in applications such as robot vision, autonomous driving \cite{liu2022petr}, and unmanned aerial vehicle system (UAV) \cite{huang2022ufpmp}. Over the past several years, Convolutional neural network (CNN) \cite{lecun1998gradient} based-object detection models have significantly improved the performance of average precision (AP) detection accuracy. However, small object detection is still a challenging task \cite{oksuz2020imbalance}. The state-of-the-art models have been reported for detecting small objects. Usually, average precision on small objects (AP$_{S}$) is relatively lower than AP on medium (AP$_{M}$) and large objects (AP$_{L}$). As MS COCO definition \cite{lin2014microsoft}, object is classified as “small” if area of segmentation mask is lower than 32x32 pixels. 

\begin{figure}[t]
\centering
\includegraphics[width=0.95\columnwidth]{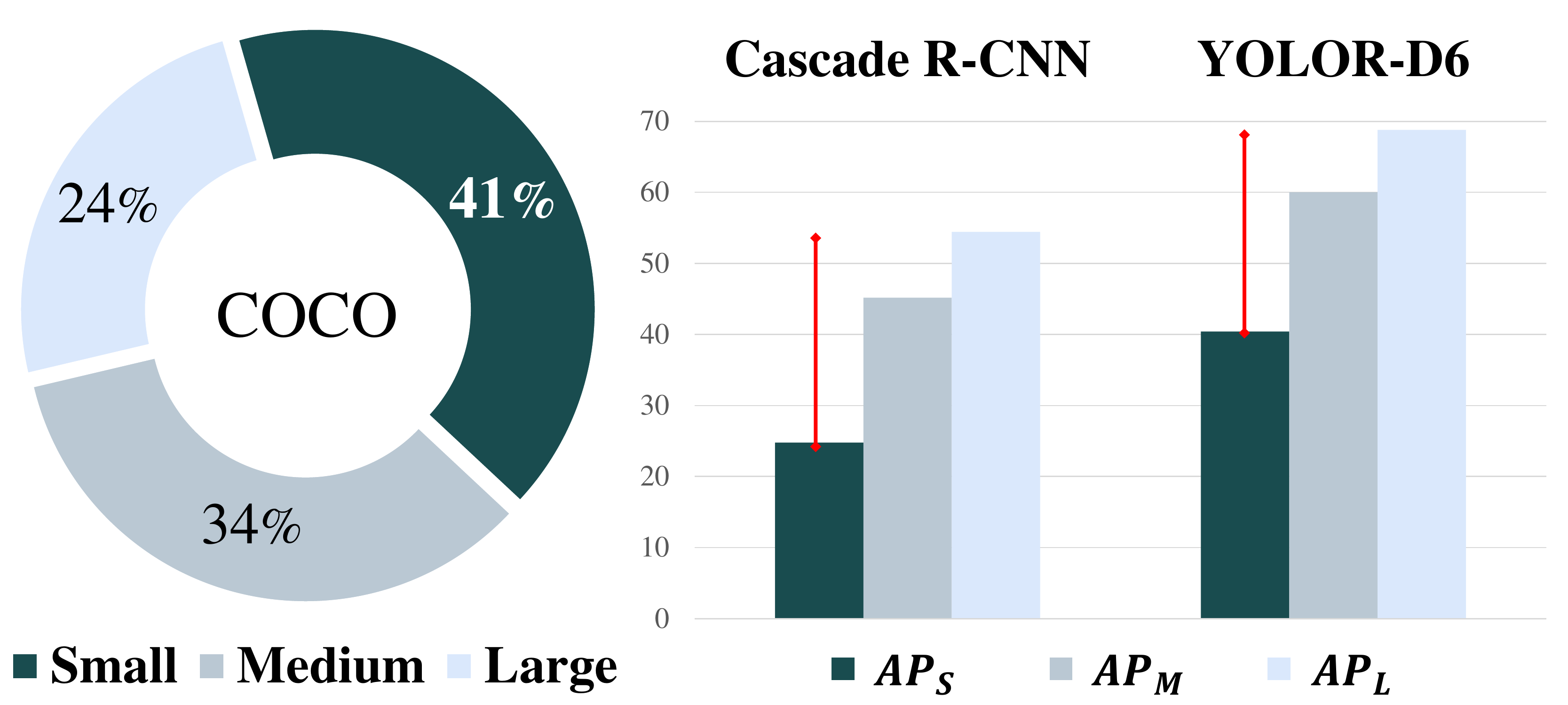} 
\caption{The proportion of object scale on MS COCO dataset and the gap of average precision (AP) between small, medium, and large scale objects on MS COCO validation set.}
\label{fig1}
\end{figure}

Figure \ref{fig1} shows the proportion of object scale and the performance gap of AP between small, medium, and large scale on MS COCO dataset. We can see small objects have the largest proportion. However, average precision on small objects (AP$_{S}$) is the lowest among other scales. Also, red line shows the performance gap from other scales.

An object has various scales in natural images, so object detection models have to be learned multi-scale features. To deal with multi-scale, scale-invariant feature has been studied in traditional computer vision \cite{lowe1999object}. The scale-invariant feature is detectable even though object scale changes. If a model learns scale-invariant feature, small object detection problem can be solved efficiently. 
Scale-space \cite{ lindeberg2013scale}, that is a multi-scale representation, is parameterized by variance of Gaussian kernel to extract scale-invariant feature. Multi-scale representation can be composed of different resolutions of images. On the other hand, recently deep learning based object detection models have used feature pyramid network (FPN) \cite{lin2017feature} as neck module to handle multi-scale objects effectively. Before detecting objects in head, they are assigned to one single pyramid level according to their scale. For example, large objects are detected in low-resolution pyramid feature map and small objects are detected in high-resolution pyramid feature map. 

To improve the performance of FPN, FPN-based models have been proposed to alleviate a semantic gap between each level pyramid feature map \cite{liu2018path}. However, most of the models are simply fusion operations like concatenation. Therefore, they could not consider the correlation of all pyramid feature maps enough.

FPN is composed of output feature maps through each convolution layer when input image is fed into CNN. The resolution of pyramid feature maps becomes smaller in process of convolution. This FPN architecture is similar to scale-space and level axis of FPN can be considered as scale axis. Therefore, scale-invariant from FPN can be extracted as in \cite{wang2020scale}. This approach motivates us to propose a scale sequence (S$^2$) feature of FPN. The higher pyramid level, the smaller image size, but semantic information is enhanced. We consider level axis of FPN as time axis of sequence and extract spatio-temporal feature by 3D convolution \cite{tran2015learning}. As a result, scale sequence feature can be a unique feature of scale-space and it is scale-invariant feature. Furthermore, All of FPN feature maps can be participated in operation using 3D covnolution. It includes a scale-correlation between all pyramid feature maps.

In comparison other scale, the reason of small object problem is that the deeper layer of CNN lead to information loss like small object feature and localization information for bounding box \cite{tong2020recent}. For small objects, we design scale sequence (S$^2$) feature built on high-resolution pyramid feature map. Generally, small objects are detected in high-resolution pyramid feature map. Therefore, we resize each pyramid feature to high-resolution feature map equally. Pyramid feature maps with extended resolution are similar to Gaussian pyramid. They are concatenated to 4D tensor for 3D convolution. This cube feature can be considered as general view referenced in Dynamic head \cite{dai2021dynamic}. After extraction, the designed scale sequence (S$^2$) feature is concatenated to high-resolution pyramid feature map for detecting small object in head.

Our contributions are three-fold: 
\begin{itemize}
    \item We propose a new scale sequence (S$^2$) feature which is extracted by 3D convolution on the level of FPN. It is scale-invariant feature of FPN regarded as scale-space. Also, all pyramid feature maps participated in operation to extract scale sequence feature.
    \item Scale sequence feature can improve AP on small objects as well as AP on other scales since built on high-resolution feature map to strengthen feature of small objects.
    \item Scale sequence feature can be extended to most object detection models based on FPN. We experimented one-stage and two stage detectors with scale sequence feature. As a result, we can observe the improved AP.
\end{itemize}

\section{Related Works}

\begin{figure*}[ht!]
\centering
\includegraphics[width=0.9\textwidth]{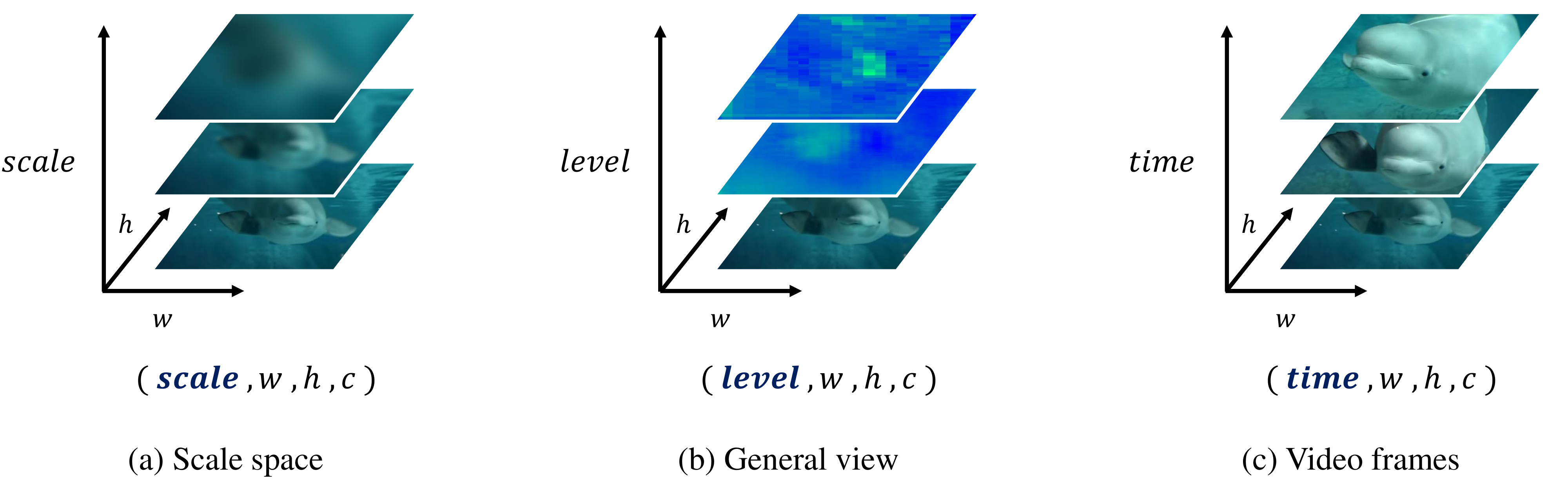} 
\caption{Comparison of image sequence on each different axes. The source of images is YouTube-8M dataset \cite{abu2016youtube}. (a) Scale space using Gaussian filter on the scale axis. (b) General view is concatenated with identical resolution features on the level axis \cite{dai2021dynamic}. (c) Video frames on the time axis.}
\label{fig2}
\end{figure*}

\subsection{Object detection models}
Object detection models have been improved along with the growth of CNN. Generally, object detection models are divided into one-stage and two-stage detector depending on the presence of a region proposal phase. Two-stage detectors extract the region of interest (RoI) from image in advance. Faster R-CNN \cite{ren2015faster} first proposed region proposal network (RPN). In Mask R-CNN \cite{he2017mask}, they added segmentation mask loss to Faster R-CNN. It proposed RoIAlign method to include more exact localization information. Also, Cascade R-CNN \cite{cai2018cascade} was reported as a multi-stage detector trained with increasing IoU thresholds. 

On the other hand, one-stage detectors conduct classification and bounding box regression simultaneously without RPN. YOLO \cite{redmon2016you} series are well-known as one-stage detectors. Scaled-YOLOv4 \cite{wang2021scaled} has proposed a scaling method of YOLOv4 \cite{bochkovskiy2020yolov4}. It has various sub-models such as YOLOv4-P5 and YOLOv4-P6 depending on the pyramid level. Recently, YOLOR \cite{wang2021you} which is state-of-the-art architecture improved the performance by unifying implicit knowledge and explicit knowledge.

The proposed scale sequence (S$^2$) feature can be applied to most object detection models. We verify the performance on both one-stage detectors and two-stage detectors with the proposed scale sequence (S$^2$) feature.
 
\subsection{Scale-invariant feature}
Scale-invariant feature \cite{lowe1999object} is defined as an unchangeable feature even though object scale changes. In traditional computer vision, scale-invariant feature has been studied to deal with multi-scale objects. Image pyramid which is a basic approach can represent various scales of objects. Also, scale invariant feature transform (SIFT) \cite{lowe2004distinctive} extracted scale-invariant feature from scale-space generated by Gaussian filters. 

Meanwhile, some research considered scale correlation in feature pyramid instead of image pyramid to reduce computation complexity. Deep scale relationship network (DSRN) \cite{wang2019dsrn} fused feature maps by bi-directional convolution. Also, pyramid convolution (PConv) \cite{wang2020scale} considered feature pyramid as scale-space and extract scale-invariant feature. Three convolutional kernels have been used for each different size feature maps and output convolutional features were added after resizing the same size.

However, these approaches compute convolution each pyramid feature map independently. In this work, we regard FPN as scale-space and extract a scale-invariant feature by 3D convolution. We defined this feature as a scale sequence (S$^2$) feature that is a unique feature of FPN. All pyramid feature maps are computed by 3D convolution. Through this process, the correlation across all pyramid features can be considered. 
Furthermore, the proposed scale sequence (S$^2$) feature includes sequence information of scale transform.

\subsection{Feature fusion strategy}
Feature Pyramid Network (FPN) \cite{lin2017feature} has been an essential module for handling multi-scale features. FPN has different resolutions of feature pyramid to assign objects according to their scales. These feature pyramids are fused by a top-down pathway. But there is a discrepancy problem between each pyramid feature map because they are generated from different depths of convolution layers. Path aggregation network (PANet) \cite{liu2018path} has been proposed for a new fusion method to alleviate the problem by adding a bottom-up pathway to FPN. 

NAS-FPN \cite{ghiasi2019fpn} found effective feature fusion strategies by AutoML training. Also, bidirectional feature pyramid network (BiFPN) \cite{tan2020efficientdet} pointed out other models considered all pyramid feature map equally regardless of their resolutions and proposed weighted fusion method for feature pyramid. Recently, Dynamic head \cite{dai2021dynamic} was introduced by using scale-aware attention that trained the importance of pyramid level adaptive to input.

However, most previous researches fused pyramid features by simply sum and concatenation. This simple structure can not consider the correlation between all pyramid feature maps. In this paper, we concatenate the proposed scale sequence (S$^2$) feature to pyramid feature maps. It reflects correlation across the whole of the feature pyramid. Therefore, it can enrich FPN for detecting multi-scale objects.

\section{Proposed Method}

\begin{figure*}[ht!]
\centering
\includegraphics[width=0.95\textwidth]{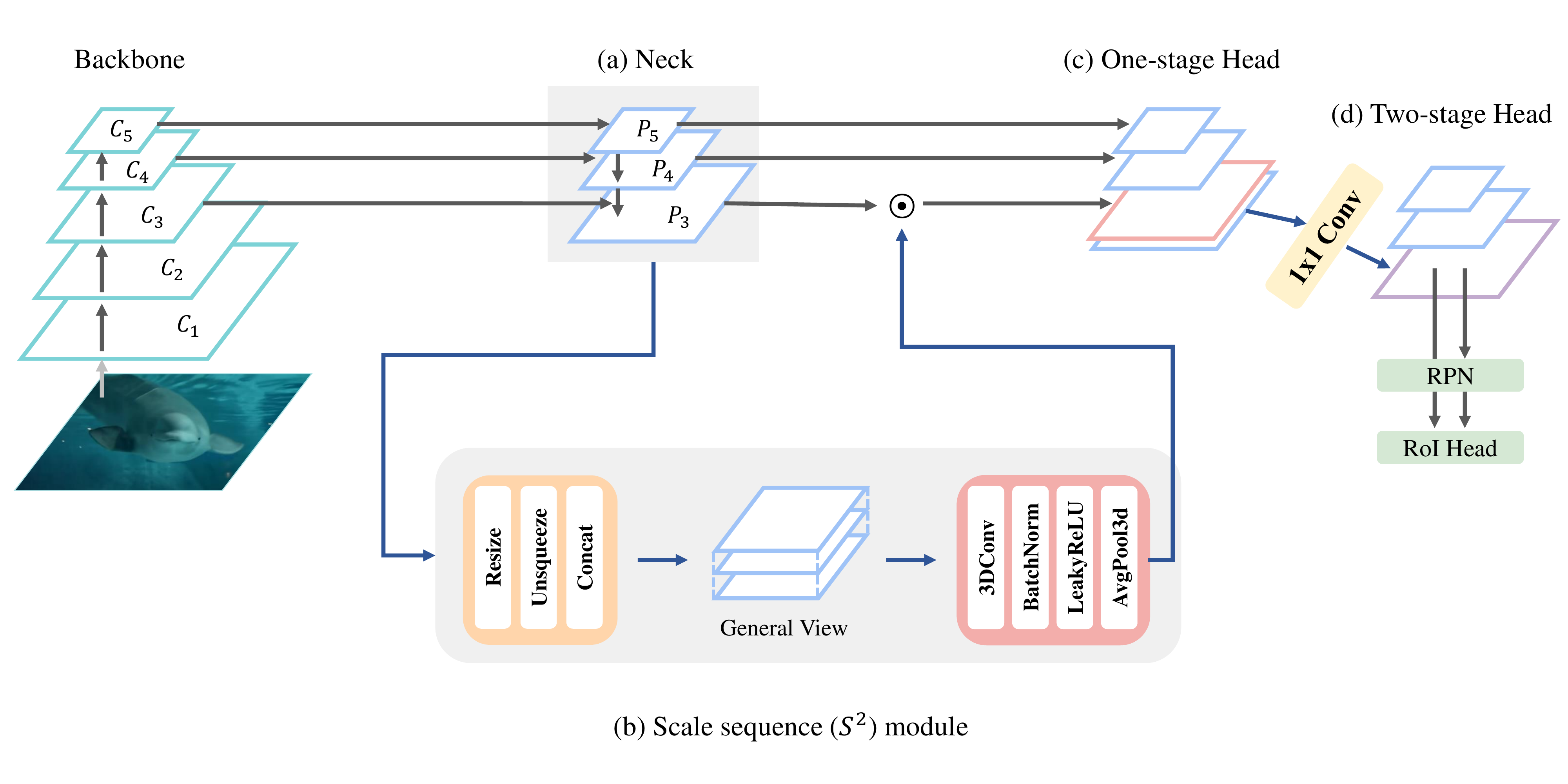} 
\caption{Scale sequence module framework: (a) Neck module of FPN for feautre fusion, (b) Process of proposed scale sequence module using 3D convolution, (c) One-stage detector head, (d) Two-stage detector head.}
\label{fig3}
\end{figure*}

\subsection{Scale Sequence (S$^2$) Feature}
In this section, we introduce a new feature: scale sequence (S$^2$). We aim to find a scale-invariant feature of FPN. The scale-invariant feature does not change although the size of image is changed. First, we explain scale-space theory \cite{lindeberg2013scale} in traditional computer vision. Scale-space is constructed along the scale axis of image. It represents not one scale, but various scale ranges that object can have. The space is generated by blurring image using Gaussian filter instead of resizing image directly. Scale-space is illustrated in Figure \ref{fig2} (a). The larger scale parameter value, the more blurred image is generated. In this theory, scale means the detailness of image. In other words, blurred image loses detail, but structural feature of the image is prominent. It is computed as follows:

\begin{equation}
g_\sigma(x,y) = \frac{1}{2\pi\sigma^2}e^{-(x^2+y^2)/2\sigma^2},
\label{eq1}
\end{equation}

\begin{equation}
f_\sigma(x,y) = g_\sigma(x,y)*f(x,y),
\label{eq2}
\end{equation}
where $f(x,y)$ is 2D image and $f_\sigma(x,y)$ is generated by smoothing through a series of convolution with 2D Gaussian filter $g_\sigma(x,y)$. $\sigma$ is scale parameter as standard deviation of 2D Gaussian filter, used in convolution. As a result, these images are same resolution but have different scale parameter values.

We consider Feature Pyramid Network (FPN) as scale-space. FPN is composed of output feature maps through each convolution layer when input image is fed into CNN. Low-level pyramid feature map is high-resolution and has information for localization, especially for small objects. On the other hand, high-level pyramid feature map is low-resolution, but it has plenty of semantic features. This property is similar to scale-space which has trade-off information on the scale axis. For based on this structure, we refer to general view from Dynamic head \cite{dai2021dynamic} that is concatenated with all pyramid features after resizing them same resolution. General view is illustrated in Figure \ref{fig2} (b). It shows feature representations are different as the level axis. Finally, we extract a unique feature of this general view from the scale view of FPN.

\begin{equation}
\mathcal{G} =  \{P_i\}_{i=3}^{L},
\end{equation}
where $P_i$ is pyramid feature map from the $i-th$ different level. The highest resolution feature pyramid is $P_{3}$. General view $\mathcal{G}$ is generated by concatenating same resolution feature maps after resizing pyramid feature map to a specific resolution. General view is made as 4D tensor: $\mathcal{G} = (level \cdot width \cdot height \cdot channel)$.

A unique feature of FPN has to consider all general view feature maps. We are motivated from 3D convolution \cite{tran2015learning} in video recognition task. In this area, 3D convolution is used to extract motion in video. Figure \ref{fig2} (c) shows video frames on the time axis. Motion is sequence as well as spatial information of frames. We regard pyramid feature maps of general view as video frames that is why general view is a sequence of convolution. The time axis of video frames can be considered level axis of general view. 

We define a unique feature of general view as scale sequence (S$^2$) feature. It is extracted by 3D convolution on the level axis of general view. This scale sequence feature is a spatio-temporal feature of general view like motion. Furthermore, all pyramid feature maps of FPN contribute 3D convolution operation. 

As a result, it can include scale correlation across feature pyramids. It is different from other FPN-based feature fusion methods that simply sum and concatenate between pyramid feature maps. The definition of scale sequence (S$^2$) feature is as the following:

\begin{equation}
S^2 feature = \Theta_{s^2}(\mathcal{G}),
\end{equation}
where $\Theta_{s^2}$ is scale sequence module based on 3D convolution. This module can extract scale sequence feature (S$^2$) from general view. For applying 3D convolution, we regard the level axis of general view as time axis of video frames as $\mathcal{G} = (time \cdot width \cdot height \cdot channel)$. As time is the length of frames, it can be denoted as the number of level of general view.

\subsection{Framework Based on Scale Sequence (S$^2$) Module}

\begin{table*}[ht!]
\centering
\resizebox{1\textwidth}{!}{
\begin{tabular}{lcc|ccc|cccc}
\hline
Model & Backbone & Size & AP & AP$_{50}$ & AP$_{75}$ & AP$_{S}$ & AP$_{M}$ & AP$_{L}$ \\
\hline
YOLOv4-P5 & CSP-P5 & 896 &  51.4 & 69.9 & 56.3 & 33.1 & 55.4 & 62.4 \\
YOLOv4-P6 & CSP-P6 & 1280 &  54.3 & 72.3 & 59.5 & 36.6 & 58.2 & 65.5 \\
YOLOR-P6 & CSPdarknet53 & 1280 & 52.6 & 70.6 & 57.6 & 34.7 & 56.6 & 64.2 \\
YOLOR-W6 & CSPdarknet53 & 1280 & 54.1& 72.0& 59.2& 36.3& 57.9& 66.1 \\
YOLOR-D6 & CSPdarknet53  & 1280 & 55.3  & 73.3 & 60.6 & 38.0 & 59.2 & 67.1 \\
\hline
YOLOv4-P5 + S$^2$ & CSP-P5 & 896 &  52.3[+0.9] & 70.7[+0.8] & 57.4[+1.1] & 34.2[+1.1] & 56.2[+0.8] & 63.7[+1.3] \\
YOLOv4-P6 + S$^2$ & CSP-P6 & 1280 &  54.8[+0.5] & 72.8[+0.5] & 60.0[+0.5] & 37.7[+1.1] & 58.5[+0.3] & 65.9[+0.4] \\
YOLOR-D6 + S$^2$ & CSPdarknet53 & 1280 &55.4[+0.1] & 73.5[+0.2] & 60.0[+0.0] & 38.1[+0.1] & 58.9[-0.3] & 67.2[+0.1] \\
\hline

\end{tabular}}
\caption{Comparison of the one-stage detectors with scale sequence (S$^2$) feature and baseline models evaluated on COCO test-dev.}
\label{table1}
\end{table*}

\begin{table*}[ht!]
\centering
\resizebox{1\textwidth}{!}{
\begin{tabular}{lc|cccc|ccc}
\hline
Model& Backbone  &AP &AP mask & AP$_{50}$ & AP$_{75}$ & AP$_{S}$ & AP$_{M}$ & AP$_{L}$ \\
\hline
Faster R-CNN & ResNet-50 & 37.9 & - & 58.1 & 41.3 & 22.0 & 40.9 & 49.1 \\
Mask R-CNN & ResNet-50 & 38.5 & 35.1 & 58.7 & 42.0 & 22.4 & 41.4 & 49.9 \\
Cascade R-CNN& ResNet-50 & 41.9 & 36.5 & 59.6 & 45.4 & 24.8 & 45.2 & 54.4 \\
\hline
Faster R-CNN + S$^2$& ResNet-50  & 39.1[+1.2] & - & 59.6[+1.4] & 42.5[+1.2] & 23.3[+1.2] & 43.0[+2.0]& 50.9[+1.7] \\
Mask R-CNN + S$^2$ & ResNet-50 & 39.8[+1.3]& 36.2[+1.1]& 60.0[+1.3] & 43.6[+1.6] & 23.5[+1.1] & 43.0[+1.6] & 51.1[+1.2] \\
Cascade R-CNN + S$^2$ & ResNet-50 & 43.2[+1.3] & 37.5[+1.0] & 60.8[+1.3] & 47.1[+1.7] & 25.8[+1.0] & 46.6[+1.4] & 56.4[+2.0]\\
\hline

\end{tabular}}
\caption{Comparison of the two-stage with scale sequence (S$^2$) feature and baseline models evaluated on COCO validation set. For comparsion, all two-stage detectors are re-trained during 3x training schedule using of 8 bath size.}
\label{table2}
\end{table*}

In this section, we explain $\Theta_{s^2}$ that is scale sequence module. Figure \ref{fig3} shows the proposed scale sequence module framework. Generally, object detection model is composed of backbone network, neck module for featuare fusion, and detection head. Input image is fed into backbone network. CNN or Transformer \cite{liu2021swin} are employed as backbone to extract feature. Convolution features through each convolution layer are denoted as \{$C_{1}$, $C_{2}$, $C_{3}$, $C_{4}$, $C_{5}$\}. Next, convolution features are aggregated by top-down and bottom-up fusion in Neck. We adopt path aggregation network (PAN) architecture instead of FPN for effective multi-scale feature fusion. Pyramid features are denoted as \{$P_{3}$,$P_{4}$,$P_{5}$\}. Figure \ref{fig3} (b) shows pyramid features fed into scale sequence module. 

In scale sequence module, scale sequence feature is designed based on $ P_{3}$ because small objects are detected in high-resolution feature map $ P_{3}$. We resize all pyramid feature maps to resolution of $ P_{3}$. To construct a general view, we add level dimension to each feature using unsqueeze function and concatenate them. This general view is fed into 3D convolution block. 3D convolution block is composed of 3D convolution, 3D batch normalization, and Leaky ReLU \cite{xu2015empirical} activation function. To reduce complexity, we employ one 3D convolution block. For small object detection, both scale sequence (S$^2$) feature and $ P_{3}$ are combined or used in detection head together. Output features from 3D convolution block are computed by average pooling 3D on level axis. Finally, scale sequence feature has identical width, height, and channel of $ P_{3}$. The new detection head for small objects has the same resolution but twice channel as:

\begin{equation}
{P_{S^2_{3}}} = CAT(P_{3} , S^2 feature),
\end{equation}
where $P_{S^2_{3}}$ is the result of concatenation between scale sequence (S$^2$) feature and $ P_{3}$ which is the highest resolution among pyramid feature maps. As a result, small objects are detected in this new detection head, $P_{S^2_{3}}$. 

We used $P_{3}$ to extract the proposed scale sequence feature for small objects by default. However, the basis resolution size for the scale sequence feature does not need to be high-resolution. It can be changed to different resolution depending on the purpose of application. 

Scale sequence module can be applied to both one-stage and two-stage detectors. Figure \ref{fig3} (c) shows the process of one-stage detector head and Figure \ref{fig3} (d) shows two-stage detector head. In order to modularize two-stage RoI head effectively, 1x1 convolution was added to $P_{S^2_{3}}$. As a result, channel size of $P_{S^2_{3}}$ of two-stage detector is identical to $P_{3}$ channel size.

\section{Experiments}

\subsection{Dataset and Evaluation Metrics}
All experiments are conducted on MS COCO 2017 dataset \cite{lin2014microsoft}. This is commonly used as benchmark dataset for object detection task. It has 80 object categories and consists of 118k train set, 5k validation set, and 20k test-dev set. We trained models on train set without extra data. Evaluation is conducted on the validation set or test-dev set by uploading our model on the official evaluation server. 

All results are evaluated by MS COCO average precision (AP). We averaged over multiple intersection over union (IoU) values. The primary challenge metric is AP at IoU=.50:.05:.95 and others are denoted as AP$_{50}$ at IoU=.50, AP$_{75}$ at IoU=.75. Also, we reported AP on different object scales. It is split into small AP$_{S}$, medium AP$_{M}$, and large AP$_{L}$ based on area measured by segmentation mask area.

\subsection{Implementation Details}
We conducted experiments to check on the performance improvement when scale sequence (S$^2$) feature is built on baseline models. For comparison, we set same training strategy and default setting that each baseline model used in their papers. We implemented all experiments using PyTorch and pre-trained COCO weights are used for initial weight. 

When adding scale sequence (S$^2$) feature, we utilize Neck modules owned by each baseline model. For example, one-stage detector, YOLO used path aggregation network (PAN) as Neck module. We implemented scale sequence (S$^2$) feature based on PAN. Also, two-stage detectors of detectron2 used feature pyramid network (FPN). Therefore, we equipped scale sequence (S$^2$) feature with FPN. All training is performed with by single-scale training without ensemble.  

For one-stage detector, we used Scaled-YOLOv4-P5, Scaled-YOLOv4-P6, and YOLOR-D6 as baseline models. Hyper-parameters and initial training options follow the setting in \cite{wang2021scaled}. We employed Stochastic Gradient Decent (SGD) as optimizer and learning rate scheduler was OneCycleLR \cite{he2019bag} with initial learning rate 0.01. One-stage detectors are trained on 3 Tesla NVIDIA V100 GPUs. Batch sizes of YOLOv4-P5, YOLOv4-P5 and YOLOR-D6 are 24, 21, and 18 respectively. The performance of one-stage detector is evaluated on MS COCO test-dev.

On the other hand, we trained two-stage detectors based on detectron2. Faster R-CNN, Mask R-CNN, and Cascade R-CNN with ResNet-50 \cite{he2016deep} backbone were selected as baseline. Three models uses FPN as Neck module. Two-stage detectors are evaluated on MS COCO validation set. For comparison, we re-trained the models using batch size 8. After built the scale sequence (S$^2$) feature, we trained two-stage detectors + S$^2$ using same batch size. Train strategy and hyper-parameters are set by detectron2’s default configuration. Training epoch is 3x scheduled (270k iteration) and learning rate is decreased by 0.1 factor at 210k and 250k iterations. Two-stage detectors were trained on 4 NVIDIA RTX 2080Ti GPUs.

\subsection{Main Results}
\subsubsection{Overall performance analysis}
\begin{table}[t]
\centering
\resizebox{1\columnwidth}{!}{
\begin{tabular}{lcccccc}
    \hline
    Level &AP & AP$_{S}$ & AP$_{M}$ & AP$_{L}$ \\
    \hline
    YOLOv4-P5& 51.4 & 33.1 & 55.4 & 62.4\\
    \hline
    P3 +S$^2$& 52.3 & 34.2[+1.1] & 56.2[+0.8]  & 63.7[+1.3]  \\
    P3,P4 +S$^2$& 52.2 & 34.2[+1.1]  & 56.4[+1.0]  & 63.3[+0.9]  \\
    P3,P4,P5 +S$^2$& 52.2 & 33.7[+0.6]  & 56.5[+1.1]  & 63.2[+0.8]  \\
    \hline
\end{tabular}}
\caption{Ablation study on different position of pyramid level for concatenating scale sequence (S$^2$) feature.}
\label{table3}
\end{table}

\begin{table}[t]
\centering
\resizebox{.95\columnwidth}{!}{
\begin{tabular}{lcccccccc}
    \hline
    Model &AP & AP$_{50}$ & AP$_{75}$ & AP$_{S}$ & AP$_{M}$ & AP$_{L}$ \\
    \hline
    YOLOv4-P5\\
    \hline
    w PAN & 51.4 & 69.9 & 56.3 & 33.1 & 55.4 & 62.4 \\
    w FPN+S$^2$ & 47.5 & 67.6 & 51.7 & 35.1 & 53.0 & 50.4\\
    w PAN+S$^2$ & 52.3 & 70.7 & 57.4 & 34.2 & 56.2 & 63.7 \\
    \hline
\end{tabular}}
\caption{Ablation study on the different Neck module.}
\label{table4}
\end{table}

We evaluated one-stage detectors built-in scale sequence feature with other YOLO-based models on MS COCO test-dev. The results are shown in Table \ref{table1}. All models with scale sequence features consistently improved the performance. For YOLOv4-P5 with scale sequence (S$^2$) feature achieved 52.3 AP which was 0.9 of improvement higher than the proposed without feature. Both AP$_{50}$ and AP$_{75}$ are improved by factor of 0.8 and 1.1 AP respectively. Also, the performance of AP is increased even if the model size is larger. For example, YOLOv4-P6 equipped with scale sequence feature has 0.5 improvement on all AP, AP$_{50}$ and AP$_{75}$. Also, YOLOR-D6 which has state-of-the-art architecture achieved 55.4 AP by applying the proposed scale sequence (S$^2$) feature.

Furthermore, we compared our scale sequence feature with two-stage object detector baselines. These experiments are evaluated on MS COCO validation set. Unlike one-stage detectors, scale sequence module for two-stage added 1x1 convolution to modularize RoI head effectively. We evaluated the proposed scheme on Faster R-CNN, Mask R-CNN, and Cascade R-CNN. 

As shown in Table \ref{table2}, Faster R-CNN with scale sequence feature achieves 39.1 AP which is 1.2 AP higher than Faster R-CNN without the proposed feature. Also, AP$_{50}$ and AP$_{75}$ have been improved by 1.4 and 1.2 AP respectively. 

For Mask R-CNN and Cascade R-CNN, the proposed scale sequence (S$^2$) feature improved the performance of AP as well as AP$_{mask}$. Mask R-CNN with scale sequence feature was improved by factor of 1.3 on AP and 1.1 on AP$_{mask}$. Also, Cascade R-CNN achieved 43.2 AP and 37.5 AP$_{mask}$ which were 1.3 and 1.0 higher than without our feature. The proposed scheme also improved on AP$_{50}$ and AP$_{75}$, significantly.

\subsubsection{Analysis of the performance as object scale}
We analyzed AP improvement on object scales: small, medium, and large. All YOLO-based one-stage detectors with scale sequence (S$^2$) feature have increased the performance of AP on all scales. In particular, AP$_{S}$ increased the most than other scales relatively. This is because the scale sequence feature was designed based on high-resolution pyramid feature map for small objects. YOLOv4-P5 which uses the smallest resolution as input image, there were 1.1 and 1.3 AP improvement in AP$_{L}$ as well as AP$_{S}$. 

On the other hand, YOLOv4-P6 has more complexity and uses larger input resolutions than YOLOv4-P5. When the scale sequence (S$^2$) feature was added to YOLOv4-P6, there was the highest improvement by 1.1 in AP$_{S}$, followed by 0.3 on AP$_{M}$ and 0.4 on AP$_{L}$ improvements. Also, YOLOR-D6 with the scale sequence (S$^2$) feature improved primary AP. It increased AP$_{S}$ and AP$_{L}$, but AP$_{M}$ was decreased slightly. 

Furthermore, two-stage detectors equipped with the scale sequence (S$^2$) feature can improve the performance of AP on all scales consistently. But the highest improvement among AP on scales is different. Because the proposed scale sequence (S$^2$) feature is added to P$_{3}$ through 1x1 convolution to adjust feature channel in two-stage scale sequence module. It caused to lack of information in P$_{3}$ compared to one-stage detectors.

\subsection{Ablation Study}
\subsubsection{Ablation study on different position of pyramid level}

We took ablation experiments on the number of scale sequence features and different positions of pyramid level. Table \ref{table3} shows the result. By default setting, the proposed scale sequence (S$^2$) feature is generated based on P$_{3}$ and has same resolution of P$_{3}$. We resized this feature to other pyramid resolutions, P$_{4}$ and P$_{5}$. Finally, the scale sequence feature is concatenated to each P$_{4}$ or P$_{5}$ and both. 

As a result, models with the scale sequence (S$^2$) feature improved their performance than those without the proposed feature. When concatenating scale sequence feature to only P$_{3}$, we achieved the best performance. It improved AP$_{S}$ as well as AP of other scales with low complexity.

\subsubsection{Ablation study on Neck model}

To analyze the effect on different Neck, we changed Neck module of Scaled YOLOv4 from PAN to FPN. Table \ref{table4} shows the result of this ablation study. We extracted a scale sequence feature from FPN instead of PAN. As a result, model with scale sequence (S$^2$) feature which is generated from PAN has better performance. 

Path aggregation Network (PAN) has two direction aggregation paths of FPN. It connects feature pyramids through top-down and bottom-up pathways. It makes all feature pyramids reflect for each other but they only have different feature map size on the level axis. This feature pyramid resembles scale-space. Therefore, it can be easily extracted scale-invariant feature from scale-space.

\begin{table}[t]
\centering
\resizebox{.95\columnwidth}{!}{
\begin{tabular}{lccccc}
    \hline
    Model & Size &Param(M). & AP & AP$_{50}$ & Speed(ms)\\
    \hline
    YOLOv4-P5 & 896 & 71 & 51.4 & 69.9& 11.8\\
    YOLOv4-P6 & 1280 & 128 & 54.3 & 72.3& 23.7\\
    YOLOR-D6 & 1280 & 152 & 55.3 & 73.3& 27.3\\
    \hline
    YOLOv4-P5+S$^2$ & 896 & 73 & 52.3 &70.7& 13.5\\
    YOLOv4-P6+S$^2$ & 1280 & 130 & 54.8 & 72.8&28.4\\
    YOLOR-D6+S$^2$ & 1280 & 155 & 55.4 & 73.5&36.2\\
    \hline
\end{tabular}}
\caption{Comparison of runtime analysis }
\label{table5}
\end{table}

\subsection{Runtime Analysis}

As shown in Table \ref{table5}, we analyzed the number of model parameters and inference speed when adding the proposed scale sequence (S$^2$) feature. Models with the scale sequence feature have increased parameters by approximately 2M. Also, we tested runtime with batch size 8 on NVIDIA Tesla V100. The speed of YOLOv4-P5 with the scale sequence (S$^2$) feature was increased by 1.7ms using 896x896 of input image. When using 1280x1280 of image as input, the speed is increased by 4.7ms and 8.9ms of YOLOv4-P6 and YOLOR-D6, respectively with the scale sequence feature. As a result, the proposed S$^2$ feature does not make large complexity in runtime.

\section{Conclusion}
In this paper, we proposed a new scale sequence (S$^2$) feature for improved object detection. It is extracted from Neck module of object detection models like FPN. The proposed feature can enrich FPN feature by reflecting a sequence of convolution that has not been considered before. Especially, the proposed feature was designed based on high-resolution pyramid feature maps for improving small object detection. It achieved improvement of AP on small as well as other scales. Scale sequence feature can be simply extended to most object detection models with FPN. Also, we demonstrated that both one-stage and two-stage detectors with scale sequence feature increased AP on MS COCO dataset.


\bibliography{aaai23}
\end{document}